\documentclass{article}
\usepackage{spconf,amsmath,graphicx}

\usepackage{hyperref}
\usepackage{xspace}

\newcommand{\pyxis}{\textsc{Pyxis}\xspace}

\title{Pyxis: an open-source performance dataset of sparse accelerators}
\name{Linghao Song, Yuze Chi, and Jason Cong\thanks{This dataset is publicly available at \url{https://github.com/linghaosong/Pyxis}.}\vspace{-9pt}}
\address{University of California, Los Angeles\\
{\tt \small \{linghaosong,chiyuze,cong\}@cs.ucla.edu}}
\renewenvironment{thebibliography}[1]{%
   \begin{oldthebibliography}{#1}%
     \setlength{\itemsep}{-.4ex}%
}%
{%
   \end{oldthebibliography}%
}

\begin{document}
\maketitle

\begin{abstract}

Customized accelerators provide gains of performance
and efficiency in specific domains of applications.
Sparse data structures and/or representations exist in
a wide range of applications. However, it is challenging
to design accelerators for sparse applications because
no architecture or performance-level analytic models 
are able to fully capture the spectrum of the sparse
data. Accelerator researchers rely on real execution 
to get precise feedback for their designs. In this work,
we present \pyxis, a performance dataset for customized
accelerators on sparse data. \pyxis collects accelerator
designs and real execution performance statistics.
Currently, there are 73.8 K instances in \pyxis. \pyxis
is open-source, and we are constantly growing \pyxis
with new accelerator designs and performance statistics.
\pyxis can be a benefit to
researchers in the fields of 
accelerator, architecture, performance,
algorithm and many related topics.

\end{abstract}

\section{Introduction}
\label{sec:intro}

Conventional processors (CPUs) are gradually 
becoming inefficient for  processing of
big-data applications
because of the diminishing gain of 
Moore's Law~\cite{moore1998cramming}.
Domain specific accelerators~\cite{cong2014accelerator,cong2019customizable,hennessy2019new,dally2020domain}
benefiting from the customization in architectures,
hardware components,
and even with the integration of 
the customization in compilers and the whole computing stack,
have been explored for higher system performance and 
energy efficiency
in many application domains such as 
deep learning~\cite{wang2021autosa,cong2018polysa,zhang2015optimizing,chen2014dadiannao,chen2016eyeriss,jouppi2017datacenter}

Because
a few parameters can determine the
whole execution of a dense workload,
it is relatively easy to use an analytic
method to model the
performance of accelerators for dense 
workload and thus guide the design
of dense accelerators. 
For example, 
we can use as few as six parameters 
to describe the processing of a convolutional layer.
Thus, Zhang et al.~\cite{zhang2015optimizing} build a
roofline~\cite{williams2009roofline} based
model for the design space exploration
of deep learning accelerators, and
Wang et al.~\cite{wang2021autosa}
employ a polyhedral model
for automatic systolic array compilation
for dense applications.

\begin{figure}[tb]
\centering
\includegraphics[width=0.85\columnwidth]{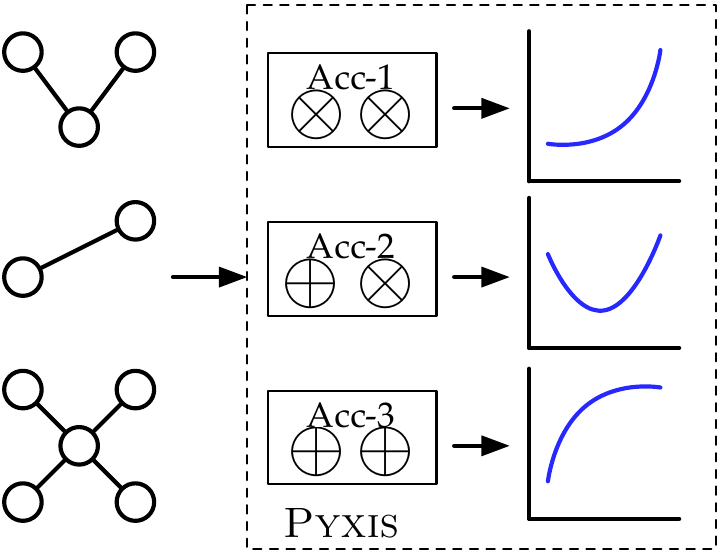}
\vspace{-9pt}
\caption{\pyxis collects the performance statistics and the corresponding accelerator designs.}
\label{figure:overview}
\vspace{-12pt}
\end{figure}

Sparse structures such as sparse matrices and graphs
encode the properties and connections, representing
data from nature, society, and a general
perspective. Besides the computation-related
optimizations, accelerator design~\cite{ham2016graphicionado,ahn2015scalable,dai2016fpgp,zhou2019hitgraph,song2018graphr,zhang2018graphp}
for sparse workloads faces more
challenges such as memory optimization, communication reduction,
and efficient data format.
However, it is impossible to use an
analytic model to guide
the design of accelerators for sparse applications
because we need the whole sparse structure to
determine the processing, and
a sparse structure is usually huge. Therefore,
instead of analytic models, real execution statistics
are helpful for the guidance of accelerator design.

There lacks a collection of performance statistics of 
accelerators on sparse workloads.
However, acquiring enough real execution statistics is very difficult.
Before the real execution,
an accelerator needs to be either implemented as
an application-specific integrated circuit (ASIC) chip
or prototyped as a field-programmable gate array (FPGA)
based accelerator. The ASIC implementation flow may
take several months, and the FPGA prototype flow requires
many hours to several days. 
The time-consuming accelerator implementation ﬂows have a deleterious impact on the generation of massive volumes of real execution statistics.
On top of this,
most accelerators can only
support a fixed-size application 
and are not general-purpose, which means one accelerator
design usually results in one or only a few performance
statistics.

The whole flow of running a sparse workload
consists of at least four components. They
are sparse data input, an application, 
an accelerator, and
performance statistics.
We present \pyxis, an open-source
performance dataset for customized
accelerators on sparse data.
We focus on the performance data
and the accelerator design in \pyxis,
as shown in Fig.~\ref{figure:overview}
To date, we have collected
73.8 K performance instances.
We use 2,637 out of all 2,893 sparse
matrices in
SuiteSparse~\cite{davis2011university} to evaluate
sparse accelerators. The matrices not
in use are out of memory.
We run sparse-matrix dense-matrix multiplication
on two FPGA platforms
(Xilinx Alveo U250 and U280) and
two GPU platforms (Nvidia Tesla K80
and V100) to collect the real execution
statistics, i.e., the latency and throughput.
We will keep growing \pyxis
by adding the latest accelerator designs
and performance statistics. 
We will also issue regular open
calls  to the whole
community to enrich the 
collections in \pyxis.

\section{Related Works}
\label{sec:related}
There are a few benchmarks 
for the evaluation of
computing systems and customized accelerators.
SPEC CPU Benchmark~\cite{henning2006spec}
is the most widely used benchmark suite
for evaluating CPU performance and
CPU architecture.
Rodinia~\cite{che2009rodinia} is a 
benchmark suite for multi-core CPUs and GPUs.
MLPerf~\cite{reddi2020mlperf} provides
a collection of machine learning applications
for system evaluation.
For the evaluation of 
high-level synthesis (HLS) based FPGA
accelerators,
Rodinia-hls~\cite{cong2018understanding}
provides the HLS versions of
Rodinia applications, and
Rosetta~\cite{zhou2018rosetta}
includes the latest deep learning applications.
For sparse data collections,
SuiteSparse
\cite{davis2011university} is the widely used
collection that contains more than two
thousand sparse matrices (graphs) from a wide
range of application domains.
The above-mentioned benchmarks/collections
mainly focus on the applications and sparse
structures. 
However, there is a lack of 
 a performance
collection of customized accelerators
for sparse workloads.

\section{Pyxis Dataset}
\label{sec:dataset}
\subsection{Method}
\noindent {\bf Sparse input data and applications.}
\pyxis collects sparse accelerators and
performance statistics.
We use 2,637 out of all 2,893 sparse
matrices in
SuiteSparse~\cite{davis2011university} 
as the sparse input data.
The sparse application we use is 
sparse-matrix dense-matrix multiplication (SpMM).
SpMM computes
$\mathbf{C} = \alpha\mathbf{A}\times\mathbf{B} + \beta\mathbf{C}$. where $\mathbf{A}$ is a sparse matrix,
$\mathbf{B}$ and $\mathbf{C}$ are dense
matrices, and $\alpha$ and $\beta$ are two
constant scalars.
We select SpMM because it is a widely used sparse
application,
and the configuration of one SpMM
$(M, K, N)$ is flexible. An input sparse matrix/graph
determines the $M$ and $K$,
but users select the $N$ value
for their applications. So we can change the $N$ value from 8 to 512
to generate seven different application settings for each sparse input.

\begin{table}[tb]
\caption{Specifications of four platforms.}
  \label{table:platforms}
  \vspace{0pt}
  \centering
  \begin{tabular}{cccc}
    \hline
    & Technology & Frequency & Bandwidth \\
    \hline
    \textbf{Alveo U250}      & 16 nm & 190 MHz   & 77 GB/s \\
    \textbf{Alveo U280} & 16 nm & 197 MHz   & 460 GB/s  \\
    \textbf{Tesla K80}    & 28 nm & 562 MHz   & 480 GB/s  \\
    \textbf{Tesla V100}   & 12 nm & 1297 MHz & 900 GB/s  \\
    
    \hline
  \end{tabular}
  \vspace{0pt}
  
\end{table}

\begin{figure}[tb]
\centering
\includegraphics[width=0.975\columnwidth]{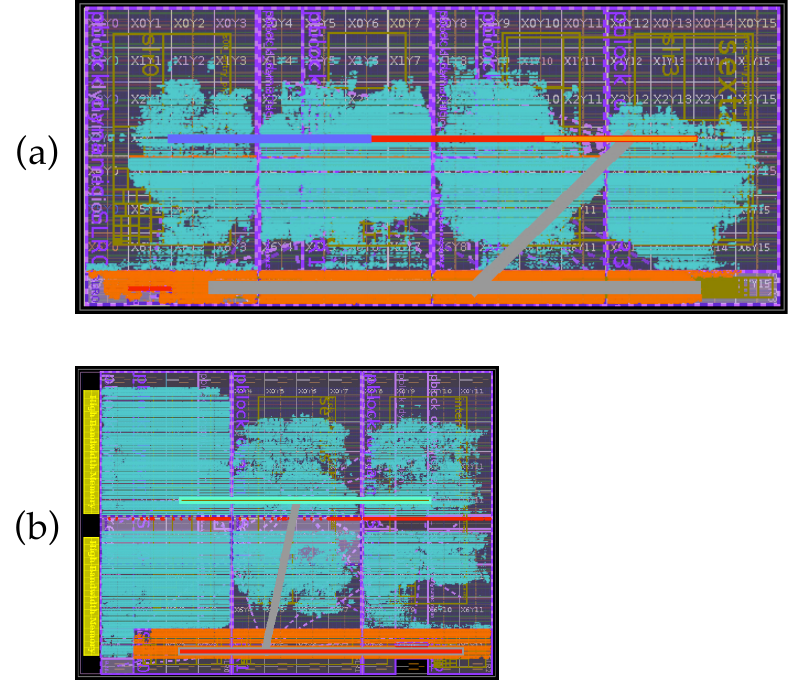}
\vspace{-9pt}
\caption{Accelerator layouts on (a) Alveo U250 FPGA and (b)~Alveo U280 FPGA.}
\label{figure:layout}
\vspace{-12pt}
\end{figure}

\begin{figure*}[tb]
\centering
\includegraphics[width=1.8\columnwidth]{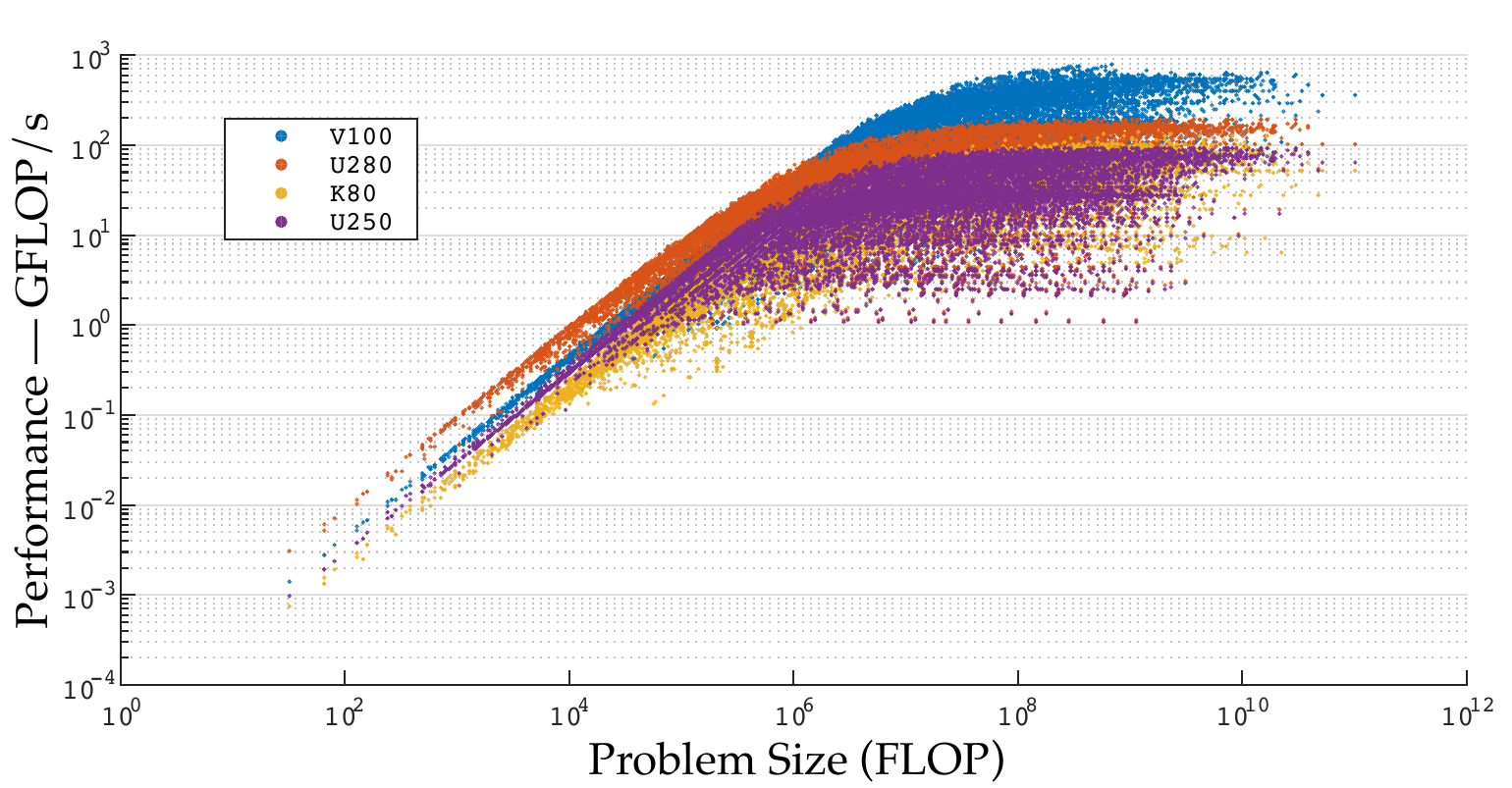}
\vspace{-9pt}
\caption{Throughputs (in GFLOP/s) v.s. problem size (in FLOP).}
\vspace{-6pt}
\label{figure:perf}
\end{figure*}

\vspace{6pt}
\noindent {\bf Sparse accelerators.}
As previously discussed, 
one accelerator design is usually for a fixed-size
application. To support a different 
application or application size, we need to run the
time-consuming accelerator prototype/manufacture flow.
Thanks to recent advances~\cite{hu2021graphlily,song2021sextans} 
in accelerator design, 
Sextans~\cite{song2021sextans}
and GraphLily~\cite{hu2021graphlily}
support an arbitrary SpMM with only one hardware
prototype generated.
We use the Sextans architecture in this work.
The accelerators are described in HLS C/C++ code. 
To generate the accelerators, we use the high level
synthesis tool Vitis 2020.2. 

We use two Xilinx FPGA accelerator cards, Alveo U250
and Alveo U280, for the accelerators. The two FPGA
cards have different memory bandwidths and 
different amounts of
logic resources. 
The Alveo U250
FPGA is equipped with a 77 GB/s DRAM, while
the Alveo U280
FPGA is equipped with a 460 GB/s high bandwidth memory (HBM). There are four super logic regions (SLRs)
on an Alveo U250 but three SLRs on an Alveo U280.
We use the two platforms to evaluate accelerator
designs with different hardware constraints.
Fig.~\ref{figure:layout} shows the layouts of the generated hardware. Besides bandwidth, the frequency 
is another factor affecting the performance of
accelerators. The layout (placing and routing)
affects the frequency.
Listed in Table~\ref{table:platforms}, Alveo U250 is 190 MHz, whereas
the frequency of Alveo U280 is 197 MHz.
Readers who are interested in
the accelerator 
architecture can find more details in~\cite{song2021sextans}.

We include the performance on GPUs because
the GPU is another type of accelerator
architecture, i.e., single instruction
multiple thread (SIMT) architecture.
We use two GPUs, Tesla K80 and Tesla V100.
In addition,
we employ CuSPARSE routine {\tt csrmm} to run
SpMMs with CUDA 10.2. Table~\ref{table:platforms}
illustrates the frequency and memory bandwidth
of the two GPU platforms.
We also include performance 
statistics from an 
Intel Xeon Gold 6244 Processor
with 16 threads and 180 GB memory.

\subsection{Performance Data}
For each SpMM run,
we collect two performance statistics: 
(1) the latency in millisecond (ms) and
(2) the processing throughput in
giga floating-point operations per second
(GFLOP/s). One SpMM run generates one data 
sample. In total, we collected
73.8 K instances in \pyxis.
Each data sample contains the
two performance statistics
and the specification of 
the sparse matrix and the SpMM.

Fig.~\ref{figure:perf} illustrates the
performance scatters of the four
platforms. We define the problem 
size as the number of floating
point operations for each SpMM.
The performance ranges from
$10^{-3}$ GFLOP/s to $10^{3}$ 
GFLOP/s.
The problem size spans a 
vast range from
$10^1$ to $10^{11}$ FLOP.
We see that
\pyxis contains rich and wide
data distribution for both
performance 
statistics and the problem size.
The rich and wide distribution
can provide sufficient information
for the following researchers who
use \pyxis dataset. 
For example, users who run an SpMM
with a problem size smaller than 
$10^{6}$ FLOP will definitely select
the U280 FPGA because of higher performance.
For hardware accelerator researchers,
it is worthwhile to explore why
SIMT architectures get higher 
performance than FPGAs on a large-size
problem.

\begin{figure}[tb]
\centering
\includegraphics[width=0.82\columnwidth]{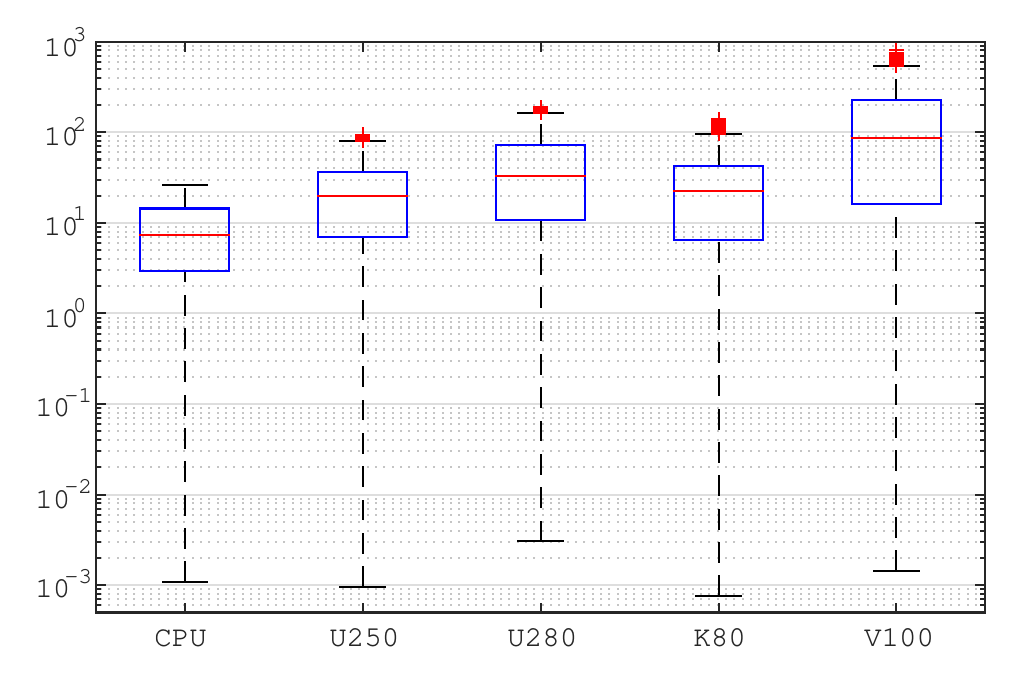}
\vspace{-9pt}
\caption{Performance distribution (GFLOP/s).}
\label{figure:boxplot}
\vspace{-12pt}
\end{figure}

Fig.~\ref{figure:boxplot} presents the 
box plots of performance on four platforms. 
The peak throughput of the CPU,
Alveo U250, U280, Tesla K80, V100 is
25.95 GFLOP/s,
94 GFLOP/s, 191 GFLOP/s, 
139 GFLOP/s, 800 GFLOP/s,
respectively.
The standard deviation of performance
of the five platforms is
6.74,
22.4, 45.0, 23.4, 152.1,
respectively.
For each platform, the performance
also covers a wide range. Thus, 
the performance information 
can help architecture 
researchers explore 
the behaviors of the different architectures
 on 
various sparse workloads.

\section{Potential Applications}
\label{sec:app}
We discuss three potential applications which can benefit from the \pyxis dataset.

\vspace{6pt}
\noindent {\bf Accelerator design.}
Although customized architectures can 
improve the performance and efficiency for 
the processing of specific applications, 
it is nontrivial to design an accelerator.
The systolic array architectures are suitable for
accelerating dense tensor 
multiplication, but there are
many versions of 
systolic array architectures. 
The first generation of Google's TPU~\cite{jouppi2017datacenter} employs
an output reuse dataflow
for the systolic array.
In the academic systolic array 
architecture Eyeriss~\cite{chen2016eyeriss}, 
Chen et al. identify three existing
data flows, i.e., weight stationary, output stationary, and no local reuse,
and present a new data flow called
row stationary. These data flow designs
are all empirical. Later, Cong and Wang~\cite{wang2021autosa,cong2018polysa}
use a polyhedral model to automatically design
the dataflows according to the size of
the dense tensor multiplication.
We see that
the research for an optimized
systolic array requires the effort from
many researchers for many years.
It is even more difficult for accelerator 
design for sparse workloads.
We hope \pyxis can provide insights,
such as bandwidth and processing unit utilization,
for
accelerator researchers.
Besides manual architectural optimizations,
automatic design space exploration~\cite{sohrabizadeh2020autodse,wu2021ironman}
and a general automatic architecture search
are significantly efficient for
accelerator design. The automatic search 
employs AI-based methods. For example,
\cite{wu2021ironman,sohrabizadeh2021gnn}
use GNNs
for optimization. AI based searching 
relies heavily
on training samples to achieve a better 
result. The performance data in
\pyxis is helpful for the
training of searching models in
automatic accelerator design.

\vspace{6pt}
\noindent {\bf Performance prediction on a new accelerator.}
Evaluating a new accelerator
takes a long time where the bottleneck is the prototype or the
manufacture flows. A fast and 
accurate prediction will save time
for accelerator researchers.
The accelerator collections in
\pyxis are essential for
performance prediction.
Currently, we do not see a
model which can take
an accelerator design as input and
predict the performance results
fast and accurately. One of the
challenges is how to encode the
accelerator design.
However, besides GNNs,
the advance in
natural language processing (NLP) models~\cite{vaswani2017attention}
can be helpful.
We describe accelerators in 
HLS C/C++ codes. NLP models
can be powerful tool to encode
the code files of accelerators.
We believe an NLP-based performance mode can provide fast and accurate feedback by 
learning from numerous accelerators. 
However, the performance model
requires further efforts from
accelerator researchers and NLP/algorithm researchers.

\vspace{6pt}
\noindent {\bf Task offloading and choice of computing platforms.}
Modern computing systems consist
of various processing units,
including CPUs, GPUs, and accelerators.
It is a challenge for users
to efﬁciently ofﬂoad the computing tasks to the available processing units.
The performance data in \pyxis
can guide the choice of the
computing platform for a specific task. For example, if a user 
runs multiple SpMMs on
a cluster equipped with a V100 
GPU and a U280 FPGA and the user 
plans to run an 
SpMM where the problem size is $10^4$ FLOP, she
is likely to run it on the FPGA
suggested by the performance 
data in Fig.~\ref{figure:perf}.
However, if the problem size is 
$10^{10}$ FLOP, she will likely run it on the V100 GPU.

\noindent {\bf Large-size graph classification/regression.}
Besides the algorithm researchers
focusing on the interdisciplinary topics of accelerator and algorithm,
\pyxis can also benefit pure 
machine learning application
researchers.
\pyxis provides meaningful labels
for many sparse matrices and graphs.
Graph classification/regression~\cite{bai2019simgnn,qin2020ghashing}
focus on the classification/regression on 
the graph level. Current graph classification/regression works on
small-size graphs where the node
number is a few hundred.
One challenge that prevents
graph classification/regression from
working on large-size graphs is
the lack of labels. Although existing
datasets such as SuiteSparse~\cite{davis2011university} contain many large-size graphs
where the node number of a graph
is up to one million,
the labels for graphs are only ID
number or datatype but not
meaningful properties.
\pyxis labels the large-size graphs
with performance metrics.
We believe that
\pyxis can help the algorithm 
advance in the
large-size graph classification/regression.

\section{Conclusion and Future Work}
\label{sec:conc}
We present \pyxis, a performance dataset for customized
accelerators running on sparse data. \pyxis collects accelerator
designs described in HLS C/C++ codes
and real execution performance statistics (latency and throughput).
Currently, there are 73.8 K instances in \pyxis. 
\pyxis can benefit  researchers in the fields of 
accelerator, architecture, performance,
algorithm, and other topics
for 
potential applications,
including accelerator design,
performance prediction, 
task off-loading, and
graph classification/regression.

\pyxis
is open-source, and we are 
constantly incorporating new accelerator designs and performance statistics to \pyxis.
Our future works include:

\noindent
{\bf $\bullet$} Incorporating more accelerator designs and performance data to \pyxis. We
are 
developing and evaluating
accelerators for 
many other sparse applications, 
such as sparse-matrix vector multiplication and graph neural
networks.

\noindent
{\bf $\bullet$} Adding more labels.
Besides latency and throughput, we
plan to add more labels, such
as power and energy consumption.
The labor (hours) for designing an accelerator
is another attractive label we are considering. 

\noindent
{\bf $\bullet$} Issuing calls for
accelerators and results. We will call for contributions from the whole community to 
contribute to \pyxis. Our standard is
open-source and reproducible.

\noindent
{\bf $\bullet$} Enriching the accelerator descriptions. 
We currently accept HLS C/C++
codes as the description of
accelerators. 
However, we will consider
accepting accelerator descriptions
in other formats.

\vspace{6pt}
\noindent{\bf  ACKNOWLEDGMENT}

This work is supported in part
by the NSF RTML Program (CCF-1937599),
CDSC industrial partners\footnote{\url{https://cdsc.ucla.edu/partners}},
and the Xilinx XACC Program.


\let\oldbibliography\thebibliography
\renewcommand{\thebibliography}[1]{\oldbibliography{#1}
\setlength{\itemsep}{-3pt}}

\bibliographystyle{IEEEbib.bst}
\bibliography{refs.bib}

\end{document}